\def\year{2022}\relax
\documentclass[letterpaper]{article} 

\usepackage{soul,color}

\usepackage{comment}
\usepackage{aaai22}  
\usepackage{times}  
\usepackage{helvet}  
\usepackage{courier}  
\usepackage[hyphens]{url}  
\usepackage{graphicx} 
\usepackage{natbib}  
\usepackage{caption} 
\DeclareCaptionStyle{ruled}{labelfont=normalfont,labelsep=colon,strut=off} 
\usepackage{algorithm}
\usepackage{algorithmic}

\usepackage{newfloat}
\usepackage{listings}
\floatstyle{ruled}
\newfloat{listing}{tb}{lst}{}
\floatname{listing}{Listing}
\newcommand{\cut}[1]{}

\title{A Human-Centric Perspective on Model Monitoring}

\author {
    Murtuza N Shergadwala,\textsuperscript{\rm 1}
    Himabindu Lakkaraju,\textsuperscript{\rm 2}
    Krishnaram Kenthapadi\textsuperscript{\rm 1}
}
\affiliations {
    \textsuperscript{\rm 1} Fiddler AI, Palo Alto, CA, USA\\
    \textsuperscript{\rm 2} Department of Computer Science, Harvard University, Cambridge, MA, USA\\
    murtuza@fiddler.ai, hlakkaraju@hbs.edu, krishnaram@fiddler.ai
}

\begin{document}

\maketitle

\begin{abstract}
Predictive models are increasingly used to make various consequential decisions in high-stakes domains such as healthcare, finance, and policy. It becomes critical to ensure that these models make accurate predictions, are robust to shifts in the data, do not rely on spurious features, and do not unduly discriminate against minority groups. To this end, several approaches spanning various areas such as explainability, fairness, and robustness have been proposed in recent literature. Such approaches need to be human-centered as they cater to the understanding of the models to their users. However, there is little to no research on \textit{understanding the needs and challenges in monitoring deployed machine learning (ML) models from a human-centric perspective}. To address this gap, we conducted semi-structured interviews with 13 practitioners who are experienced with deploying ML models and engaging with customers spanning domains such as financial services, healthcare, hiring, online retail, computational advertising, and conversational assistants. We identified various human-centric challenges and requirements for model monitoring in real-world applications. Specifically, we found that relevant stakeholders would want model monitoring systems to provide clear, unambiguous, and easy-to-understand insights that are readily actionable. Furthermore, our study also revealed that stakeholders desire customization of model monitoring systems to cater to domain-specific use cases. 
\end{abstract}

\section{Introduction}\label{sec:intro}
Machine learning (ML) is increasingly playing an integral role in our day-to-day experiences. Increasingly, the applications of ML are no longer limited to search and recommendation systems, such as web search and movie and product recommendations, but ML is also being used in decisions and processes that are critical for individuals, businesses, and society. With ML based solutions and pipelines in high-stakes applications such as hiring, lending, criminal justice, healthcare, and education, the resulting personal and professional implications of ML are far-reaching. Consequently, it becomes critical to ensure that the underlying ML models are making accurate predictions, are robust to shifts in the data, are not relying on spurious features, and are not unduly discriminating against minority groups. This emerging field, called {\em model monitoring}, can be viewed as part of a broader ML model governance \cite{kurshan2020towards} and responsible ML framework~\cite{arrieta2020explainable}, and is at an inflexion point, as evidenced by legal/regulatory requirements, requirements from the perspective of web-scale ML applications, and adoption of practical and scalable approaches. Model monitoring is receiving greater attention in light of regulations such as EU GDPR, CCPA, and the EU Trustworthy AI\footnote{https://ec.europa.eu/digital-single-market/en/news/ethics-guidelines-trustworthy-ai} initiative and several ML deployment failures in practice.

\begin{figure*}[hbt!]
\centering
\includegraphics[trim=0 100 0 0, clip, width=0.9\textwidth]{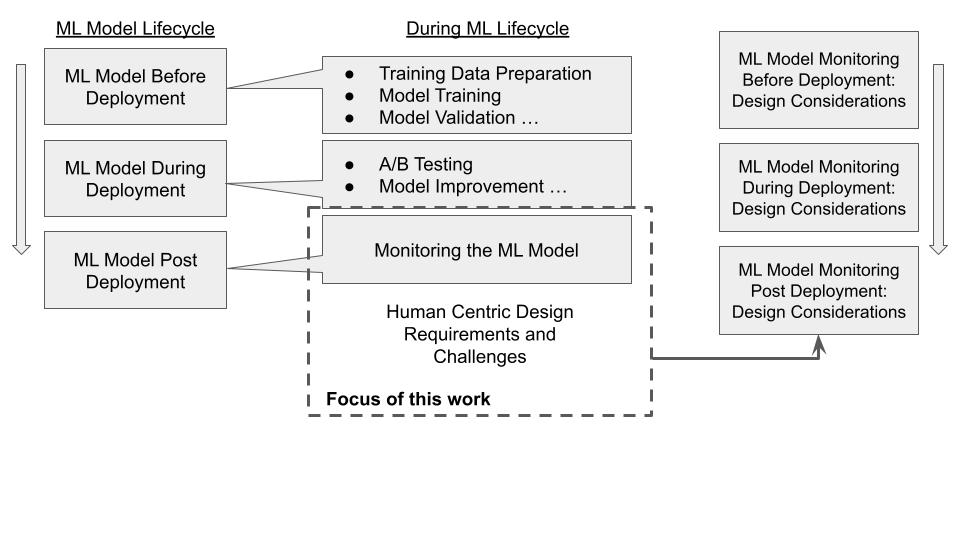}
\caption{Various activities during ML model life cycle and the focus of this work. }
\label{fig:focus}
\end{figure*}

From an operational angle, large-scale ML systems require maintenance not only by the virtue of possessing software code but also because of the nuances of ML as a domain itself~\cite{sculley2015hidden}. ML-specific nuances include dependency on data whose distributions can shift during production from when the model was designed and the dependency of a model on another model's output which can cascade issues from the other model onto the dependent model~\cite{sculley2015hidden}. Such nuances imply that maintenance of ML systems requires \textit{monitoring} its various aspects, such as data and models. Such monitoring is essential to ensure that the model does not become stale or degrade in performance due to changing real-world conditions or changes in the data collection process and data processing pipelines.
In Figure~\ref{fig:focus}, we illustrate the various activities in the lifecycle of an ML model and highlight the focus of this work in the monitoring activity.
More broadly, the emerging field of {\em model monitoring}
pertains to practices for deploying and maintaining ML models in production reliably and efficiently~\cite{makinen2021needs}. Monitoring of deployed ML models is needed to determine how often the model needs to be retrained and handle the following issues: (1) Data drift: The distribution of features may change over time, causing the quality of predictions made by the model to gradually degrade~\cite{breck2019data}. (2) Changes in the relationships between input and target variables: The changes in real-world conditions may alter the relationships between input and target variables (often referred to as ``concept drift''~\cite{gama2014survey, tsymbal2004problem}), and thereby result in degraded model performance. (3) Data integrity and operational challenges: Compared to traditional software, ML models are more tolerant to unintended or unexpected changes in their inputs, and hence may continue to make predictions, even when the inputs may have been corrupted. As a result, the predictions may be erroneous or of poor quality. Thus, it is essential to ensure data integrity and detect any undesirable changes in the data pipeline (e.g., changing the measurement unit of a feature from feet to yards). (4) Reduced performance for subgroups of users: Although the model performance may not change as a whole, it is possible for the model to exhibit poor performance for certain subgroups of users (for whom, say, the relationship between input and target variables may have changed). Hence, in addition to monitoring overall accuracy and other performance measures, it is important to ensure that the model does not develop bias, and instead performs well across various subgroups of users.


We note that the above issues are of interest not only for data scientists but also for ML engineers, product managers, business decision makers, policy, compliance, and legal teams, internal and external auditors, and other stakeholders. In other words, given the importance of model monitoring as part of a broader AI model governance framework, model monitoring need to be human-centered not just in terms of usability by humans but also accounting for human behavior~\cite{shneiderman2021responsible,wing2021trustworthy}. In this work, our goal is to understand the needs and challenges in monitoring deployed ML models from a human-centric perspective. This perspective is absolutely critical and is missing from existing literature.

{\noindent \bf Key Contributions}: The goal of our study is to unearth practical and real-world challenges that are often encountered in real-world settings employing machine learning models. We conducted semi-structured interviews with 13 practitioners who are experienced with deploying ML models and engaging with customers spanning domains such as financial services, healthcare, hiring, online retail, computational advertising, and conversational assistants. We identified various human-centric challenges and requirements for model monitoring in real-world applications. Specifically, we found that relevant stakeholders would want model monitoring systems to provide clear, unambiguous, and easy-to-understand insights that are readily actionable. Furthermore, our study also revealed that stakeholders desire customization of model monitoring systems to cater to domain-specific use cases.

\section{Related Work}\label{sec:related}
This work lies at the intersection of several emerging areas of machine learning research, namely, detecting dataset shifts, monitoring model behavior via model understanding and explanations, monitoring fairness and robustness of machine learning models, and human-centered studies and open source tools focused on the aforementioned aspects. Below, we discuss some of the key works across each of the aforementioned areas. 

\paragraph{Dataset Shifts} There is a rich literature on techniques for detecting shifts in the data (e.g., see \citet{breck2019data, cormode2021relative, gama2014survey, karnin2016optimal, tsymbal2004problem, webb2016characterizing, vzliobaite2016overview} and the references therein). 
Both verifying the validity of model inputs and detecting changes in the features or model outputs are important challenges encountered in practical ML applications. The former is often addressed by including user-defined tests such as tests to check if a feature value is within a specified range~\cite{schelter2018automating}. For the latter, statistical hypothesis testing and confidence interval based approaches have been proposed. Statistical hypothesis testing involves checking if two given sets of samples are drawn from the same distribution by using a test statistic. Student's t-test and Kolmogorov–Smirnov test are examples of commonly used tests~\cite{wasserman2004all,murphy2012machine}. More advanced tests such as Maximum Mean Discrepancy can also be used for higher dimensional data~\cite{gretton2012kernel}. As these tests require sufficient fine-tuning (e.g., selecting the kernel and its hyperparameters), confidence interval based approaches~\cite{efron1994introduction} are often employed for detecting drifts in practice~\cite{model_monitor_2021}. In addition to the above approaches which are model-agnostic, specialized methods that leverage the model internals and the training data have also been proposed to determine the extent of drift and take remedial steps~\cite{NEURIPS2020_219e0524,lipton2018detecting,reddi2015doubly,wu2019domain}.

\paragraph{Interpretability} 
As argued by several recent works~\cite{doshi2017towards}, model understanding is absolutely critical to ensure that ML models are relying on appropriate features when making predictions. To this end, model interpretations and explanations are widely being used to monitor model behavior. 
Many approaches have been proposed to directly learn interpretable models for various tasks including classification~\cite{letham15:interpretable,WangRu15,lakkaraju16:interpretable,lou2012intelligible,bien2009classification} and clustering~\cite{kim14:the-bayesian,lakkaraju16:confusions}. To this end, various classes of models such as decision trees, decision lists~\cite{letham15:interpretable}, decision sets~\cite{lakkaraju16:interpretable}, prototype (case) based models~\cite{bien2009classification}, and generalized additive models~\cite{lou2012intelligible,caruana15:intelligible} were proposed. However, complex models such as deep neural networks and random forests are often shown to achieve higher accuracy than simpler interpretable models~\cite{ribeiro16:kdd}; thus, there has been a lot of interest in constructing post hoc explanations to understand their behavior. 

A variety of post hoc explanation techniques have been proposed, which differ in their access to the complex model (i.e., black box vs. access to internals), scope of approximation (e.g., global vs. local), search technique (e.g., perturbation-based vs. gradient-based), explanation families (e.g., linear vs. non-linear), etc. For instance, LIME~\cite{ribeiro16:kdd} and SHAP~\cite{lundberg17:a-unified}, are \emph{model-agnostic}, \emph{local explanation} approaches that explain individual predictions of any black box model by training a linear model locally around each prediction.
These approaches rely on input perturbations to learn these interpretable local approximations. 
Several other \emph{local explanation} methods have been proposed that compute \emph{saliency maps} which capture importance of each feature for an individual prediction by computing the gradient with respect to the input~\cite{simonyan2013saliency, sundararajan2017axiomatic, selvaraju2017grad,smilkov2017smoothgrad}. A number of other local explanation methods~\cite{koh2017understanding,ribeiro2018anchors} have also been proposed in the literature. 
An alternate approach is to provide a global explanation summarizing the black box as a whole~\cite{lakkaraju19:faithful,bastani2017interpretability}, typically using an interpretable model. 

Some recent work has shed light on the downsides of post hoc explanation techniques. \citet{rudin2019stop} argues that post hoc explanations are not reliable, as these explanations are not necessarily faithful to the underlying models and present correlations rather than information about the original computation. 
There has also been recent work on exploring vulnerabilities of black box explanations~\cite{adebayo2018sanity,slack2019can,lakkaraju2020how,rudin2019stop,dombrowski2019explanations}---e.g., \citet{ghorbani2019interpretation} demonstrated that post hoc explanations can be unstable, changing drastically even with small perturbations to inputs. 

\paragraph{Fairness} It is crucial to ensure that ML models deployed in real-world applications do not unduly discriminate against minority subgroups. To this end, monitoring the fairness of ML models has become common place in recent times~\cite{DworkHPRZ12,HardtPS16}. 
The initial literature on fairness in machine learning emphasized heavily on outlining the precise definitions of statistical fairness~\cite{HardtPS16}. Several competing and contrasting notions of fairness emerged during this phase which can be broadly categorized into: 1) \emph{group fairness} which emphasizes that protected groups should receive similar treatment as that of advantaged groups~\cite{BerkHJKR18,HardtPS16} 2) \emph{individual fairness} which requires that \emph{similar} individuals to be treated similarly~\cite{DworkHPRZ12}, and 3) counterfactual fairness which captures the intuition that a decision pertaining to an individual is fair if it is the same in the actual world and a counterfactual world where the individual belonged to a different demographic group~\cite{kusner2017counterfactual}. Furthermore, various metrics have been proposed to realize each of the aforementioned notions of fairness. For example, statistical (demographic) parity, equalized odds, equality of opportunity, and predictive parity are metrics proposed to enforce group fairness. 

There are pros and cons to each of the aforementioned notions and metrics of fairness. For example, \citet{DworkHPRZ12} argue that the group fairness notion of statistical parity leads to highly undesirable outcomes e.g., one might end up incarcerating women who pose no safety risk to ensure the same proportions of men and women are released. On the other hand, \citet{kim2018fairness} highlight that assessing individual fairness is often hard in practice because it is hard to determine what is an appropriate metric function to measure the similarity of two individuals. Similarly, realizing counterfactual fairness in practice is also non-trivial because we do not have access to the counterfactuals of real world decisions i.e., there is no ground truth to determine if someone would have been incarcerated if that individual was a male instead of being a female and vice versa. Furthermore, prior research has also established that certain notions of fairness (calibration and balance conditions) are fundamentally incompatible and cannot be simultaneously optimized~\cite{KleinbergMR17,Chouldechova17}. 

\paragraph{Open Source and Commercial Tools} 
Several open source and commercial frameworks for monitoring deployed ML models have been developed in recent times. Examples of such frameworks include Amazon SageMaker Model Monitor \cite{model_monitor_2021} \& Clarify \cite{hardt2021amazon}, Deequ \cite{schelter2018automating}, Evidently \cite{evidently}, Fiddler's  Explainable Monitoring \cite{fiddler_monitor}, Google Vertex AI Model Monitoring \cite{google_drift}, IBM Watson OpenScale \cite{ibm_monitor}, Microsoft Azure MLOps \cite{azure_monitor}, and Uber's Michelangelo platform~\cite{uber_michelangelo}.
In contrast to these tools and techniques, the focus of this work is on understanding the needs, requirements, and challenges associated with monitoring deployed models from the perspective of relevant stakeholders. 

\paragraph{Human-Centric Perspectives and User Studies} 
There have been several user studies and interviews to understand the desiderata for model explanations and fairness~\cite{doshi2017towards,cheng2021soliciting}. For instance,~\cite{bhatt2020explainable} conducted interviews with data scientists to understand the use cases and accompanying desiderata for explaining models. On the other hand,~\citet{lakkaraju2020fool} carried out a user study to understand if misleading explanations can fool domain experts into deploying racially biased models, while \citet{kaur2020interpreting} found that explanations are often over-trusted and misused. 
Similarly, \citet{poursabzi2018manipulating} found that supposedly-interpretable models can lead to a decreased ability to detect and correct model mistakes, possibly due to information overload.
\citet{lage2019evaluation} used insights from rigorous human-subject experiments to inform the design of explanation algorithms. While the above works touch upon human-centric perspectives on explainability and fairness, they do not focus on human-centric perspectives on model monitoring which is the key goal of our work.

\section{Study Design}\label{sec:studydesign}

We collected desiderata data for model monitoring from 13 practitioners who are experienced with deploying ML models and engaging with customers spanning domains such as financial services, healthcare, hiring, online retail, computational advertising, and conversational assistants. We did so by conducting semi-structured and one-on-one interviews virtually and analyzed the results from each interviewee. All the interviewees had an understanding about ML model monitoring and had interacted with various model monitoring tools.

In the following, we discuss the experimental design choices, the data collection process and the data analysis approach for the study.

\subsection{Data Collection}

We collected desiderata from thirteen (n=13) ML practitioners with expertise in the model monitoring space. We recruited these participants by reaching out to ML practitioners who are either working on developing model monitoring tools or needing such tools for their application use cases. 

Based on an average of 2.7 years of professional experience of the interviewees within the MLOps space, we considered four to be experts, four to be beginners, and five to be intermediate. The interviews were semi-structured due to the exploratory and human-centered nature of the study which required prompting interviewees with follow-up questions relevant to their domain-specific experiences with model monitoring. The interview questions and their rationales are tabulated in Table~\ref{tab:interview_questions}.

\begin{table}[tb!]
\centering
\begin{tabular}{|p{0.22\textwidth}|p{0.2\textwidth}|}
\hline    \textbf{Question} & \textbf{Motivation}  \\ \hline
    IQ1: What kind of applications do you use ML models for? & To understand domain-specific use cases  \\
    IQ2: Why do you need model monitoring in these applications? & To understand domain-specific desiderata  \\
    IQ3: What aspects of model monitoring do you need?  & To understand interpretations of model monitoring  \\
    IQ4: What would an ideal model monitoring framework look like and what do you want this framework to tell you? & To understand human-centered desiderata for model monitoring \\ \hline
\end{tabular}
\caption{Interview Questions (IQ) for Model Monitoring in Practice}
\label{tab:interview_questions}
\end{table}

\subsection{The Choice of Subject Pool: MLOps Practitioners}
We chose ML practitioners working in the Machine Learning Operations (MLOps) space for several reasons. First, MLOps is a relatively new and emerging field where questions such as IQ3 in Table~\ref{tab:interview_questions} are still utilized to understand people's perspectives on what constitutes as model monitoring as there is no set definition for it yet. Thus, MLOps practitioners are well versed with model monitoring related challenges as they face them in their day to day activities. Second, the MLOps practitioners we interviewed have experience with domain-specific use cases and would thus be able to better articulate fuzzy application and business requirements. Third, MLOps practitioners develop model monitoring solutions which enables them to discuss practical implementation challenges as well.

\subsection{Data Analysis}\label{subsec:analysis}

We analyzed the responses to all the questions using inductive content analysis~\cite{krippendorff1980content}. The interview notes were analyzed in two ways. First, each transcript was individually annotated for model monitoring application areas/use cases, requirements, and challenges described by each of the interviewees. Second, the responses to each question mentioned in Table~\ref{tab:interview_questions} across all interviews were pooled and analyzed to inductively generate common themes and categories for ML monitoring  application areas, design requirements, and challenges.

Responses to IQ1 were identified either as a domain or a use case. For example, phrases and words such as ``financial services", ``insurance", ``banking", and ``adtech" were labeled as domains. Whereas, phrases such as ``fraud detection", ``credit lending", and ``speech recognition" were labeled as a use case. Responses to IQ2, IQ3, and IQ4 were analyzed to identify human-centric desiderata for model monitoring. Sentences that contained phrases such as ``need to", ``should have", ``be able to", and ``it would be great if" were labeled as requirements. Phrases such as ``difficulty", ``challenges", ``not possible", ``hard", and ``risky" were analyzed to identify challenges faced by the interviewees. Further, Responses to IQ3 were also analyzed to discover interviewees' interpretation of model monitoring and what it entails. To do so, authors' domain knowledge of model monitoring was leveraged to label the aspect of model monitoring discussed by an interviewee. For example, if interviewees discussed data drift as a part of the response to IQ3, data drift was labeled as an aspect of model monitoring for that interviewee. The IQ3 labels across all interviewees were pooled to characterize various aspects of model monitoring as discussed by them.

\section{Application Areas, Design Considerations and Challenges for Model Monitoring: Practitioners' Perspectives}\label{sec:desiderata}

In this section, we discuss the use cases, requirements, and challenges discussed by the interviewees. We begin with the results of IQ3 analysis, where we describe various aspects of model monitoring. Then, we discuss the results of IQ1 analysis, namely, the application areas mentioned by the interviewees. Then, we discuss the interviewees' desiderata and challenges for model monitoring.











\subsection{What is Model Monitoring? Practitioners' Perspective}\label{subsubsec:results_what}

We analyzed the responses to IQ3 contextually to identify the key aspects of model monitoring stated by the interviewees. All the interviewees discussed \textit{model performance monitoring} and \textit{data drift monitoring} as a part of model monitoring. One interviewee discussed, \textit{``we are interested in prediction drift as a proxy for model performance."} Another interviewee mentioned, \textit{``Data scientists care more about model performance, deploying more models, etc."}. Similarly, one interviewee highlights their need for model monitoring by mentioning, \textit{``we need tools for continuous model monitoring to assess end-to-end impact of these [design] changes and ensure model performance."} 

Five (out of thirteen) interviewees emphasized \textit{monitoring model fairness, bias, and model versions} as a part of model monitoring. One interviewee said, \textit{``A subset of our models needs fairness analysis to ensure compliance."} Another interviewee mentioned their interaction with a banking client as follows: \textit{``If you are a risk manager for banks, then you probably care about fairness/bias."} One interviewee discussed fairness monitoring as an aspect of model monitoring and mentioned, \textit{``Monitoring fairness could be another interesting addition for my clients who are at the receiving end of regulatory constraints, fines, etc."} With respect to model versioning, an interviewee stated, \textit{``At} [retracted organization name], \textit{ we have a champion model and a contender model. We used things such as A/B tests to evaluate both the models. [This is where] I think model monitoring can immensely help."}

We also noticed interviewees categorizing or grading the relative importance of various aspects of model monitoring as a part of the response to IQ3. Interviewees largely identify monitoring data integrity~\cite{boritz2005practitioners}, that is, the accuracy, completeness, and consistency of data for inputs and outputs of a model as a \textit{basic requirement}. One interviewee remarked, \textit{``I think, at the very least, [we need] a system that monitors the exhaust of the model; the outputs and also the inputs of the model."} In addition to this, identification of data drift, performance drift, and outlier detection were mentioned as important but \textit{intermediate requirements} for model monitoring. Another interviewee stated, \textit{``On top of basic input output monitoring, another aspect is model performance from accuracy and precision point of view."} Finally, monitoring model fairness and bias were expressed as regulation driven, ``good to have", and are thus labeled as \textit{advanced requirements} for model monitoring.

\subsection{Application Areas and Use Cases}\label{subsubsec:results_application}
Analyzing responses to IQ1 enabled us to identify specific domains and use cases of interest to the interviewees as shown in Table~\ref{tab:application_results}. Some example excerpts include: \textit{``of course, classic are fraud, churn use cases, and recommendation for anything [in] retail."} Another interviewee states, \textit{``our primary market is financial services and fintech, we have some retail, but insurance is clearly a new area where we are seeing new interest."} Regarding the financial services domain, one interviewee specifically mentioned that \textit{``they're [financial services domain] the most advanced with the use of machine learning or they have the highest bar of regulatory scrutiny."} In the context of monitoring for online advertisement, one interviewee said, \textit{``if by mistake we advertise to the wrong user, and by mistake, we bid with the wrong price, that will affect the end results, the revenues or the profit."} While there are numerous other domains and use cases for ML models, the results here are intended to contextualize the responses of the interviewees to the desiderata and the challenges discussed 
below.

\begin{table}[tb!]
\centering
\begin{tabular}{|p{0.1\textwidth}|p{0.31\textwidth}|}
   \hline
   \textbf{Domain} & \textbf{Use Case}  \\ \hline \hline 
    AdTech & Ads personalization and ads pricing. \\\hline 
    Consumer Technology & Wake-word detection, automatic speech recognition, natural language understanding and interpretation, entity resolution, and text-to-speech generation. \\\hline 
    Financial Services & Fraud detection, credit lending, and churn prediction. \\\hline 
   Insurance & Risk prediction \\\hline 
   Retail & Recommendation models and traffic monitoring. \\ \hline
\end{tabular}
\caption{Domains and use cases discussed by the interviewees.}
\label{tab:application_results}
\end{table}

\subsection{Human-centric Requirements for Model Monitoring}
The following themes for model monitoring requirements emerged based on interviewees' responses to IQ2, IQ3, and IQ4.


{\noindent \em Domain-Specific Debugging \& Root Cause Analysis}: Interviewees discussed the need for a model monitoring system to discover slices or sub-populations of data where unexpected model behavior and outcomes occur. This would help one gain insight on model errors, when to retrain a model, and  domain-specific nuances. One interviewee mentioned, \textit{``If it can provide me with some early warnings or signs where things go wrong and give me ways to resolve what's going on."} Further, the system should allow customizable levels of abstractions such as feature-level monitoring, prediction-level monitoring, and performance-level monitoring based on the use case. One interviewee stated, \textit{``the thing that would really help is customizing monitoring by allowing different overlays of time scale, business metrics, model metrics, etc."}.

{\noindent \em Risk Management, Model Governance, and Privacy}: Interviewees would like model monitoring systems to help them manage risk and ensure regulatory compliance. Interviewees emphasized the need for a monitoring system to enable centralization of model governance in an organization rather than have dependencies on an individual or a team that created the model. One of the interviewees cited the work of~\cite{kurshan2020towards} in discussing the challenges of model governance and mentioned the need for frameworks with self-regulatory capabilities. Such a system was also discussed to require to reduce human dependency and automate the process of error detection in ML pipelines. One of the interviewees remarked, \textit{``In the past, people would manually go [look for errors] and document [the errors] but now we are trying to automate by monitoring."} Interviewees discussed the risk associated with retraining the model without monitoring. One interviewee said, \textit{``[without monitoring] model refresh can be too risky. How do we ensure the new model is working as intended?"} In certain settings, it would also be desirable for the monitoring system to ensure that privacy and confidentiality of various assets such as protected user information in training data and intellectual property associated with the models are protected. One of the interviewees mentioned, \textit{``that's [privacy] keeping people from not sending the data to a hosted environment; they want to keep it in their own private VPC and that adds a lot of challenges for monitoring [as a service]." }

{\noindent \em Human-Centered Design}: Interviewees discussed the need to preserve human autonomy and decision-making. In other words, the monitoring system should not trigger actions such as automated retraining, but instead provide actionable insights and suggestions to enable humans to make better decisions. Further, the monitoring system should provide relevant and meaningful alerts without cognitive overload. One interviewee discussed, \textit{``For individual data scientists, I think the process [of root cause analysis] becomes cumbersome or if they are trying to share information, it becomes untenable when the number of models grows."} In the context of emotions, one interviewee stated, \textit{``you know if somebody wakes me up at 12 o'clock in the night saying `Oh, there is a drift and take a look,' and if there is no drift, I am going to be mad."} Another interviewee said, \textit{``An ideal monitoring system should not create alert fatigue."} To avoid this, the model monitoring system needs to be aware of aspects such as the types of alerts, how often they are fired, and how they are presented. The challenge, however, as discussed by one interviewee is that, \textit{``we may not have the right metrics that measure human [centered] things such as individual preferences and fatigue."}

\subsection{Challenges for Model Monitoring: Temporal Categorization}

We temporally categorized challenges as (1) design challenges before deploying a monitoring system, (2) challenges during monitoring an ML system, and (3) challenges post deployment with respect to the usefulness of monitoring outcomes. 

Interviewees highlighted several \textit{practical challenges in designing and deploying an ML model monitoring system}. 
These challenges included design questions such as what should be monitored, and how should the monitoring system interact with the model. One interviewee discussed, \textit{``sometimes you have models that translate from one layer to another. So monitoring a feature XYZ may not make sense to the customer."}  Another interviewee discussed that for monitoring systems as a service, \textit{``when it comes to actually doing things, there are a lot of other constraints that become more important like people [clients] might not talk about [their use case] or might not explicitly tell you so."} These point to the requirement of a monitoring system having the ability to provide some guidelines or in-built options for what to monitor in an ML system.

Interviewees also discussed whether the monitoring system would also need the technical dependencies that a model requires, such as packages and modules, to ingest a model and execute successfully. One interviewee discussed, \textit{``let's say an organization uses a package but there I'm not quite sure what the business model is, can anybody go and download that and set that up and run that on their end? I'm not sure."} Thus, monitoring systems are discussed as systems that have the ability to take an ML system as an input and provide monitoring insights as an output. In that context, interviewees raise the challenge for an ML monitoring system to possess a super-set of the technical capabilities of different ML systems.
We note that this \textit{perceived challenge} by the interviewees in practice is only a concern if the ML model is required to run specific predictions on the monitoring platform. We discuss this finding in Section~\ref{sec:discussion}.

Some interviewees had prior experience with model monitoring systems, and discussed the challenges of protecting privacy of users in training data and confidential information / intellectual property associated with their models.  One interviewee said, \textit{``For third party monitoring tools offered as SaaS [Software-as-a-Service], the clients have to send the data [to the third party server] -- that becomes a privacy challenge."}
All interviewees emphasized the challenges of adapting a monitoring system to the domain-specific needs of the ML system and the lack of solutions that cater to their specific needs. For example, the data volume experienced by an ML system is highly sensitive to its application context, and could become a key design consideration for an ML monitoring system to be able to handle. 
Such decisions are currently made manually, and there is a lack of a framework that helps automate the design process for an ML monitoring system. Hence, interviewees highlighted the need for a fully managed monitoring service for their ML systems.

We also observed that the interviewees discussed \textit{challenges that may occur when an ML monitoring system is deployed}. 
They discussed the lack of existing reliable solutions in assessing whether an observed drift in data or model performance is a cause for concern. One interviewee mentioned \textit{``products have some sort of seasonality or recurring drift that is probably not very interesting for the developers to get alerted for. For example, when I was at [organization name] we would see usage peak on weekends vs. weekdays because most people would open the app on the weekends. If [organization name] were to use a [model] monitoring system, how do we identify drift that is anomalous vs. simply seasonal [or periodic]?"}
Such a lack of reliability can result in cognitive fatigue that may desensitize practitioners from gaining meaningful insights from the monitoring system. Interviewees also discussed latency challenges, that is, how quickly can a monitoring system detect issues and suggest remedial actions. Here, interviewees discuss the need for the monitoring system to inform the human user about the computation time of various aspects of monitoring that may not necessarily be computed on similar timescale. One interviewee stated, \textit{``I can do an F1 score in real time and that's a simple thing to be able to do, given a particular classification threshold but if I want to do AUC which is independent of the classification threshold, all of a sudden that's a very different computation. So it's very interesting that people know the formula, but they expect the same behavior for both which is not theoretically possible."} Systems that are slow to identify issues may not be useful in certain contexts, such as autonomous driving, where high-stakes decisions often need to be made in real-time by ML models. Further, due to privacy, intellectual property, and security/compliance considerations, interviewees highlighted that the monitoring tools and systems may need to be present “on-premise” or in-situ rather than a third party housing such a system. Consequently, debugging or maintaining the monitoring system itself may be challenging, especially if the services are being sought from a third party.

Interviewees also discussed the \textit{``so what” or value-based challenges} with monitoring systems. These discussions focused on the value of the insights gained from such systems, and pertained to aspects such as whether/how the monitoring insights could enable the stakeholders to take concrete actions to improve business outcomes and whether non-experts would be able to understand these insights. As an example, one interviewee quoted, \textit{``developers need to know how the model affects business metrics, and monitoring is important for that."}

\subsection{Challenges for Model Monitoring: Feature-specific Categorization}
As noted earlier, data drift, outlier detection, data integrity violation, model performance, and bias/fairness 
are the key dimensions of model monitoring highlighted by the interviewees. Next, we discuss the specific desiderata themes within some of these dimensions.

For data drift, we found that all the interviewees consider input and output data distribution monitoring as a necessity for model monitoring. Also, such drift monitoring is treated as an indicator for model retraining. Further, data integrity violation and outlier detection are aspects that are discussed as a part of the data drift monitoring functionality as well. One interviewee said, \textit{``customers want simple stuff like data drift, performance monitoring, and some sort of anomaly detection. That is the first set of things they want."} Another interviewee mentioned, \textit{``customers are very worried about data integrity issues such as a breakdown in their pipeline they want to know immediately. A sudden change in data."} We note here that the ``customers" referred to by the interviewees are stakeholders interacting with these interviewees in professional settings who have expressed interest in using model monitoring systems as a service. 

For bias/fairness monitoring, customer requirements are currently driven through policies and regulations. One interviewee stated, \textit{``Fairness is rare so far in my experience but with big banks, regulation and compliance are driving it."} Another interviewee said, \textit{``There is a smaller subset of cases for bias and fairness monitoring."} Words such as ``policy",``regulatory constraints",``fines", and ``regulatory push" were used to describe the need for fairness monitoring. Practitioners mentioned that compliance and risk management teams are concerned about bias/fairness also due to its impact on the trust and the reputation of a company. By leveraging ML model monitoring, an organization could proactively detect and mitigate any biases observed in its deployed models instead of having to react when such issues are discovered by external entities.

\section{Conclusion and Discussion}\label{sec:discussion}
Motivated by the need for understanding the human-centered requirements and challenges in designing monitoring frameworks for ML systems, we performed an interview study with ML practitioners with experience spanning several application domains. We presented findings and insights on real-world use cases, desiderata, and challenges for ML model monitoring in practice based on these interviews. Interviewees discussed both feature-specific and process-specific aspects of model monitoring. Feature-specific aspects include monitoring data drift, model performance, and bias/fairness and ensuring that the alerts are relevant without cognitive overload. Process-specific aspects include the temporal considerations before, during, and after the deployment of the model monitoring system and the ability of a monitoring system to cater to different needs across the lifecycle of an ML system.

Based on the requirements and challenges reported in this work, we discuss potential pathways to address concerns on themes of human-centric design of model monitoring systems. 
We believe that human autonomy and agency can be preserved using a human-AI decision making framework, wherein the model monitoring system is used in conjunction with a human decision maker.
This implies enabling the system to report data integrity violations or different types of drift, and then allowing human users to pursue corrective pathways such as correction of labels or features, data re-sampling, and model retraining. To avoid cognitive load, monitoring systems could include threshold knobs and preference logging features that enable humans to embed domain-specific knowledge for their specific use cases. For centralization of model governance, the model monitoring system could enable automatic report generation and use of natural language to describe the state of an ML system. This would enable non-experts to also have an understanding of the state of an ML system. To address latency concerns, while there may be technological advancements to improve computation speed as well as efficient designs of a system, we also highlight the need for the system to educate the human user about the time estimates for certain computations as well as a comparative view of the difference in compute time for different aspects of a system. Such knowledge can potentially improve user experience while leveraging model monitoring systems. 


We highlight a \textit{perceived challenge} 
regarding the technical requirements of a model monitoring system. Interviewees described the challenge for an ML monitoring system to possess a super-set of the technical capabilities of different ML systems it monitors. However, we note that to monitor inputs, outputs, and other characteristics associated with a deployed model, a monitoring system does not need to execute the model and can instead take the production logs associated with the model as input. Thus, the model monitoring system may not need the technical infrastructure to run the model itself, and can instead focus on the \textit{tools and techniques for prediction of distribution shifts}. This understanding influences the design decisions for monitoring systems and we thus highlight that model monitoring systems may be designed in a model-agnostic manner. This also points to the caution required in analyzing human-centric desiderata where perceived challenges by practitioners, who may not necessarily be experts in ML, may stem from misconceptions about the system functionalities.

We also note that inferring a user's cognition is currently an active area of research~\cite{akula2022cx, shergadwala2022does,wu2022inferring}. While some human-centric requirements may point to the user's desire to be ``understood" by a system in real-time, it is currently not practically feasible to so. Further, several studies have highlighted the limits of human oversight and the challenges that arise when attempting to build tools that enable people to monitor technological systems (e.g., see \citet{perrow1999normal} and \citet{green2022flaws}). Thus, we need to be aware of the limits in the ability to build model monitoring tools and interfaces that actually satisfy the human-centric requirements stated by the interviewees.

We acknowledge that our analysis is limited based on the inputs of only thirteen practitioners. 
However, the interviewees are ML practitioners with deep knowledge of ML systems in their respective domains and extensive experience of ML model monitoring systems. Hence, we were able to obtain and analyze human-centric requirements and challenges from a practical viewpoint based on our interviews.
More broadly, we encourage MLOps practices to formalize design frameworks for ML monitoring systems that are cautiously informed by human-centered desiderata.

\section*{Acknowledgments}
We are thankful to the interviewees for their detailed responses to the questions. We also thank Joshua Rubin and Lea Genuit for their feedback and help with the analysis.

\bibliography{aaai22}

\end{document}